\documentclass[runningheads]{llncs}

\usepackage{eccv}

\usepackage{eccvabbrv}

\usepackage{graphicx}
\usepackage{booktabs}

\usepackage[accsupp]{axessibility}  

\usepackage{hyperref}

\usepackage{orcidlink}
\usepackage{bbold}
\usepackage{xcolor}
\usepackage{multirow}
\usepackage{multicol}
\usepackage{dsfont}
\usepackage{amssymb}
\usepackage{amsmath}

\definecolor{myred}{HTML}{DF7162}
\definecolor{myblue}{HTML}{6A9BDD}
\begin{document}

\title{DiffuSearch: How Hybrid Trajectory Planning Benefits from Aligned Objectives in Diffusion and Action Space} 

\titlerunning{DiffuSearch}

\author{Steffen Hagedorn\inst{1, 2}\orcidlink{0000-0003-3443-0087} \and
Aron Distelzweig\inst{1, 3}\orcidlink{0009-0003-4750-4855} \and
Alexandru P. Condurache\inst{1, 2}\orcidlink{0000-0002-0626-335X}}

\authorrunning{S. Hagedorn \etal}

\institute{Robert Bosch GmbH, Germany\\
\email{steffen.hagedorn@de.bosch.com} \and
Institute for Neuro- and Bioinformatics, University of Lübeck, Germany \and
Department of Computer Science, University of Freiburg, Germany}

\maketitle

\begin{abstract}
        In trajectory planning for autonomous driving, hybrid planning architectures are often realized as a collection of disparate modules, each with its own objectives.
        This lack of a unifying principle can lead to inconsistencies between the initial and refined trajectory, resulting in suboptimal behavior.
        We address this by introducing \texttt{DiffuSearch}, a novel hybrid planner that uses a unified set of objectives across generation and refinement.
        Our model encourages all components to follow the same shared driving goals: collision avoidance, drivable area compliance, comfort, and progress.
        \texttt{DiffuSearch} employs a two-stage architecture.
        First, a guided diffusion model generates a scene-consistent, joint trajectory prediction, using our driving objectives as differentiable guidance functions to implicitly steer the denoising process.
        Second, a Monte Carlo Tree Search (MCTS) in a discretized action space performs an explicit, local refinement of this proposal, leveraging the same driving objectives as its reward function.
        This synergistic design leverages the diffusion model's strength in finding scene-consistent solutions combined with the explainable, constraint-aware refinement of MCTS.
        Experiments on nuPlan and interPlan reactive closed-loop benchmarks demonstrate that \texttt{DiffuSearch} achieves strong and often state-of-the-art performance, substantially reducing collisions and improving comfort, particularly in complex, interactive scenarios.
        Our ablation studies indicate that MCTS refinement is the main mechanism behind the gains, while sharing objectives between implicit guidance and explicit search provides further consistent improvements.
  \keywords{Autonomous Driving \and Action Planning \and Machine Learning}
\end{abstract}

\section{Introduction}
\label{sec:intro}
\begin{figure}[!tb]
    \centering
    \includegraphics[width=\columnwidth]{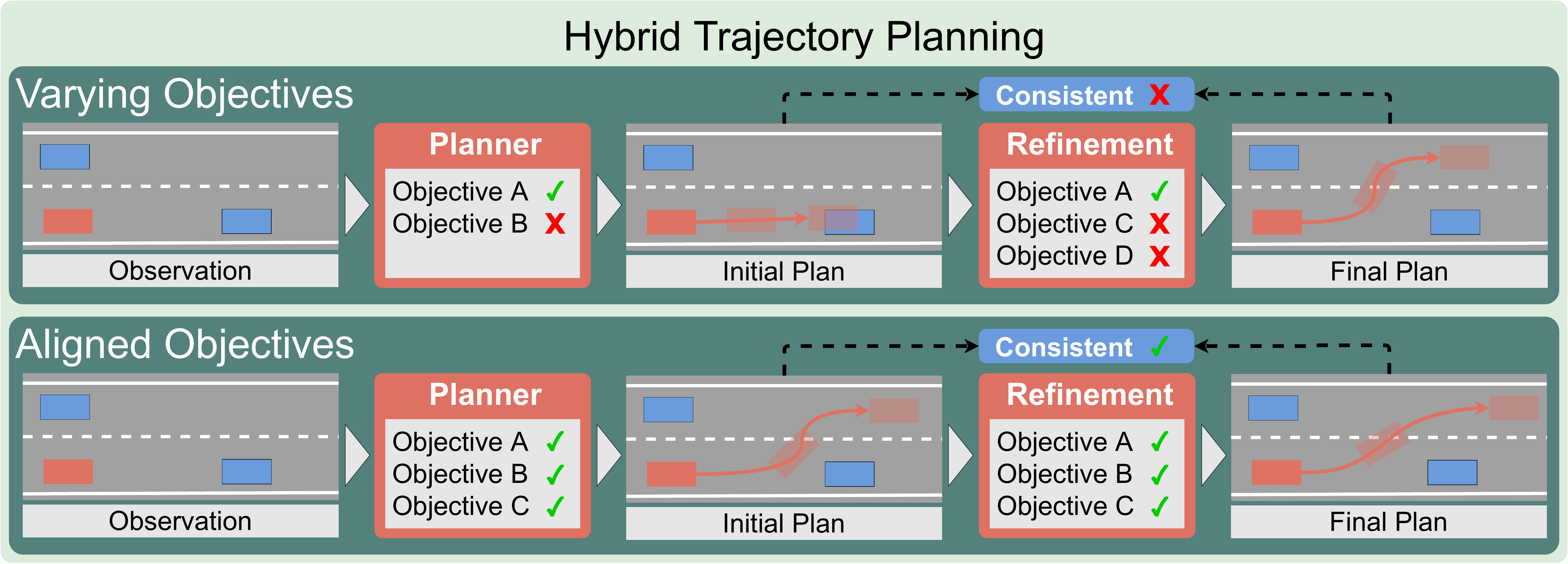}
    \caption{\textbf{Varying vs. unified driving objectives:} Current hybrid planners use varying objectives in different modules. This can lead to undesired effects, where the refinement changes the initial plan to a completely different and suboptimal behavior. We use a shared set of driving objectives across modules to facilitate a continuous refinement of the initial trajectory.}
    \label{fig:cover_figure}
\end{figure}
In autonomous driving systems, trajectory planning is the vital link between perception and actuation.
While modern deep learning-based methods offer flexible behavior in diverse traffic scenarios, they provide limited explainability and do not by themselves expose explicit safety checks that are essential for deployment.
Consequently, many state-of-the-art planners use hybrid approaches that apply a refinement stage to make the final trajectory better satisfy critical driving objectives~\cite{huang2023gameformer, hu2023planning, chen2024vadv2, cheng2024pluto, zheng2025diffusion}. 
Unfortunately, this refinement stage is often disconnected from the initial proposal generator, with each module optimizing for its own objectives (\cref{fig:cover_figure}).
This lack of a unifying principle can lead to inconsistencies where the refinement radically changes the initial plan rather than improving it, resulting in suboptimal behavior.
We address this issue by proposing a model built on unified objectives, encouraging all components to work towards the same goal to achieve a continuous refinement of the trajectory.
Further, our model is designed to meet the theoretical requirements for interactive and safe planning:
The ideal planner would repeatedly interleave prediction and planning to model bidirectional interactions between agents~\cite{hagedorn2024integration}.
However, the computational demand of this approach is prohibitive.
A strong practical alternative is a joint prediction and planning model, which can implicitly capture these bidirectional interactions within a single, unified step~\cite{bansal2018chauffeurnet, sadat2019jointly, zheng2024genad, zheng2025diffusion}.
Our method, \texttt{DiffuSearch}, extends this concept to a two-stage architecture: a joint diffusion~\cite{ho2020denoising} model backbone followed by a Monte Carlo Tree Search (MCTS) refiner.
This design combines the diffusion model's ability to capture the joint, multi-modal data distribution of a traffic scene with the MCTS's strength in performing focused, explainable optimization that explicitly evaluates constraint violations for the ego trajectory~\cite{niu2024planning, ziegler2023anytime, chekroun2024mbappe, yu2025hype}.
The diffusion model first generates an initial trajectory proposal by leveraging classifier-free guidance~\cite{ho2022classifier}, where differentiable functions representing our driving objectives (collision avoidance, drivable area compliance, comfort, and progress) implicitly steer the generation process.
However, since this implicit guidance cannot formally guarantee compliance, we use an explicit MCTS in a discretized action space to refine the plan, using the very same driving objectives as its reward function.
This unified objective provides a strong inductive bias, supports explainability, and enables explicit constraint-aware refinement.
Our contributions are threefold:
\begin{enumerate}
\item We present a novel trajectory planner, \texttt{DiffuSearch}, that combines a guided diffusion backbone with an MCTS refinement in action space, using an initial joint prediction as a strong prior to effectively confine the search space.
\item We show that MCTS refinement is the primary source of improvement, and that using the same objectives in classifier-free diffusion guidance and the MCTS reward provides additional gains.
\item We demonstrate that our model achieves competitive and state-of-the-art results in reactive closed-loop simulation across nuPlan and interPlan benchmarks with different background traffic agents.
\end{enumerate}

\section{Related Work}

\subsection{Hybrid Trajectory Planning}
Deep learning-based trajectory planning for autonomous driving has focused on different paradigms throughout the past years.
Many early methods used monolithic neural networks and imitation learning to directly regress the planned trajectory for the ego vehicle~\cite{bojarski2016end, codevilla2018end, xu2017end}. 
While flexibly adapting to diverse traffic scenarios, these methods lack explainability and safety guarantees~\cite{hagedorn2024integration}.
Consequently, modular end-to-end architectures became popular, dividing the model into interpretable modules to facilitate introspection and introduce domain-knowledge, while still optimizing the whole model for the final planning task~\cite{zeng2019end, casas2021mp3, renz2022plant}.
To reliably enforce safety-critical constraints, many models implement hybrid planning strategies that generate trajectory proposals with an (end-to-end) learned model and select or refine them based on interpretable driving objectives~\cite{dauner2023parting, huang2023gameformer, hu2023planning, chen2024vadv2}.
Typical driving objectives comprise drivable area compliance, collision avoidance, comfort, and progress along the route~\cite{dauner2023parting, sun2025generalizing, zheng2025diffusion}.
However, trajectory refinement is often based on different driving objectives than in previous modules, which can lead to suboptimal solutions.
We address this shortcoming by using a shared objective across proposal generation and refinement to improve model consistency and planning performance.

\subsection{Diffusion Models}
Denoising diffusion probabilistic models (DDPMs) are generative models that excel at capturing complex, multi-modal data distributions~\cite{ho2020denoising, dhariwal2021diffusion}.
A neural network, often a Transformer~\cite{peebles2023scalable}, is trained to iteratively denoise a sample, starting from pure Gaussian noise and conditioned on relevant context information.
This is formally achieved by learning the score function $\mathbf{s}_\theta(\mathbf{x}_t, t)$, which approximates the gradient of the log-probability of the noisy data distribution, $\mathbf{s}_\theta(\mathbf{x}_t, t) \approx \nabla_{\mathbf{x}_t} \log p(\mathbf{x}_t)$~\cite{song2020score}.
The key advantage of DDPMs for planning lies in their controllability at inference time via guidance.
Classifier-free guidance~\cite{ho2022classifier} steers generation by adding a $w$-weighted corrective term derived from a differentiable objective function $\mathcal{O}(\mathbf{x}_t)$
\begin{equation}
\label{eq:guidance}
\hat{\mathbf{s}}_{\theta}(\mathbf{x}_t, t) = \mathbf{s}_{\theta}(\mathbf{x}_t, t) - w \nabla_{\mathbf{x}_t} \mathcal{O}(\mathbf{x}_t, t).
\end{equation}
In autonomous driving, diffusion models have been applied to motion prediction~\cite{jiang2023motiondiffuser}, planning~\cite{sun2025generalizing, zheng2025diffusion}, and traffic simulation~\cite{zhong2022guided}.
Works like \texttt{Diffusion} \texttt{Planner}~\cite{zheng2025diffusion} have qualitatively shown that classifier-free guidance can generate safe and comfortable trajectories.
Notably, they have not reported improvements through guidance on scale but only exemplarily.
Moreover, implicit guidance offers no formal guarantees that the final trajectory complies with all constraints.
This motivates the need for a subsequent refinement stage, which to date is disconnected from the generative model's own objectives~\cite{sun2023large, sun2025generalizing, zheng2025diffusion}.
\texttt{DiffuSearch} addresses this disconnect by using the same driving objectives first for implicit diffusion guidance and then for explicit MCTS evaluation.

\subsection{Monte Carlo Tree Search}
Monte Carlo Tree Search (MCTS) is a heuristic search algorithm for sequential decision-making~\cite{coulom2006efficient, kocsis2006bandit, silver2016mastering}.
It incrementally builds a search tree by balancing the exploration of new action sequences with the exploitation of known high-reward paths, making it well-suited for refining an initial plan.
In autonomous driving, MCTS is applied by framing planning as a search over a tree of future vehicle states and driving actions~\cite{essalmi2025extended, wen2024monte}.
However, MCTS is fundamentally iterative and can be computationally expensive in vast search spaces without a strong prior to guide the search.
Our work leverages this synergy: we use the high-quality trajectory from our diffusion model as a strong prior, enabling the MCTS to perform a focused, local refinement rather than a slow, unguided exploration.
A key design choice in MCTS-based planning is the definition of the search space.
While some variants search in an abstract latent space~\cite{schrittwieser2020mastering}, a common approach in driving is to search in a discretized action space (typically acceleration and steering angle)~\cite{niu2024planning, ziegler2023anytime, chekroun2024mbappe}.
This provides inherent explainability, as each tree branch is an interpretable maneuver, and allows for explicit constraint enforcement via the reward function for each branch.
Recent hybrid planners have confirmed the value of using learned priors to guide this explicit search, e.g., \texttt{HYPE}~\cite{yu2025hype} uses learned ego proposals, while \texttt{MBAPPE}~\cite{chekroun2024mbappe} uses a learned world model to simulate the consequences of actions.
End-to-end differentiable tree planners such as \texttt{TPP}~\cite{chen2023tree} and \texttt{DTPP}~\cite{huang2024dtpp} optimize learned latent pipelines, but sacrifice this explicit explainability.
In contrast, no action-space MCTS planner has aligned the reward function with the objective of the learned prior generator.
\texttt{DiffuSearch} closes this gap by using the same shared driving objectives for diffusion guidance and MCTS refinement, making the search focus on a prior already shaped by similar preferences.

\section{Methodology}
\label{sec:methodology}
\begin{figure}[!tb]
    \centering
    \includegraphics[width=\columnwidth]{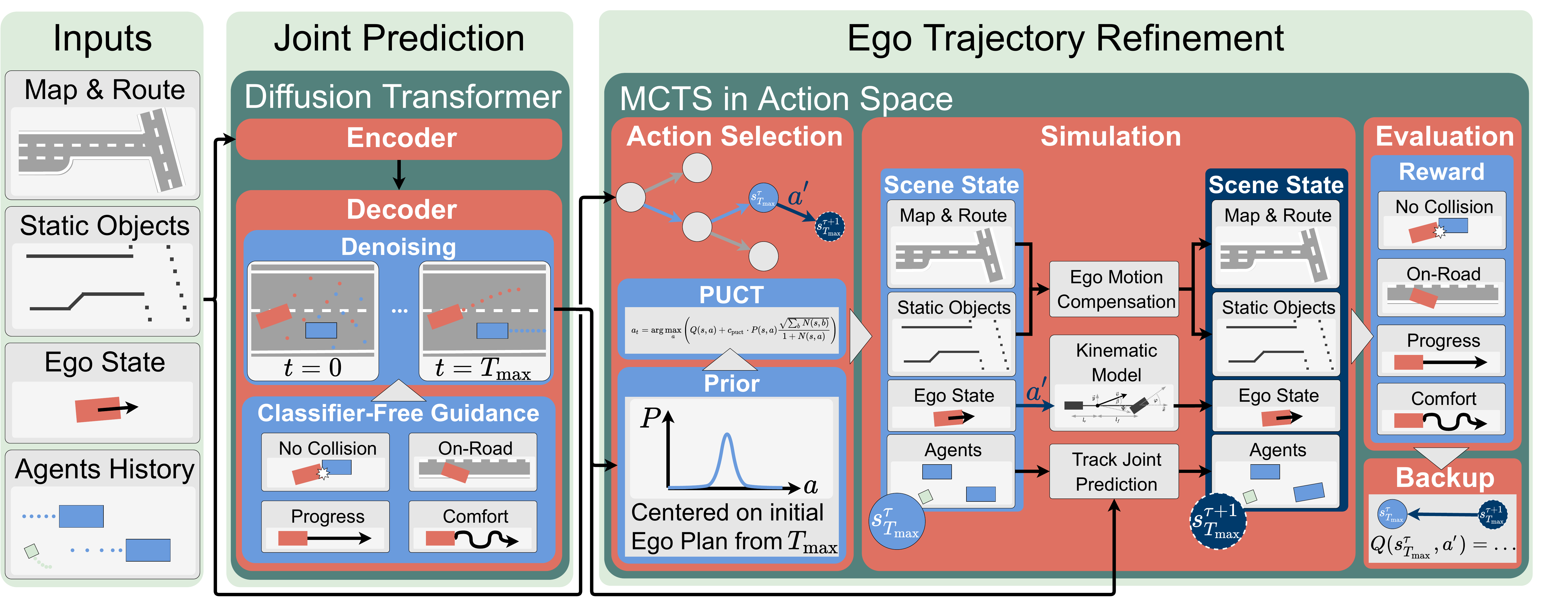}
    \caption{\textbf{\texttt{DiffuSearch} model overview.} A diffusion transformer takes object-level inputs to make a joint prediction, guided by differentiable driving objective functions. The ego trajectory of this joint prediction is used as the prior distribution for MCTS refinement. In a focused search around the initial ego plan, action sequences are selected with the PUCT rule and simulated using a kinematic model for the ego vehicle and diffusion-based joint predictions for other agents. The reward consists of the same driving objectives used to guide the diffusion model, facilitating a consistent refinement of the initial ego trajectory.}
    \label{fig:model_overview}
\end{figure}

The theoretical challenges of autonomous driving, balancing global scene understanding with local, safety-critical optimization, motivate a hybrid architectural design.
Our proposed method, \texttt{DiffuSearch}, addresses this by combining a generative diffusion model with a search-based model for explicit refinement.
This two-stage process, illustrated in \cref{fig:model_overview}, is built upon a unified set of driving objectives that guide both stages, encouraging a coherent planning pipeline.

\subsection{Joint Prediction via Guided Diffusion}
\label{sec:method_diffusion}
The first stage of our framework is a generative model tasked with generating a scene-consistent joint prediction.
This joint prediction serves as a strong prior for the subsequent refinement stage, constraining the search space to a region around this initial solution.
We formulate joint prediction as a conditional generation problem, where the goal is to jointly generate future trajectories for the ego vehicle and all surrounding agents in a fixed radius, conditioned on the current scene context (\cref{fig:model_overview}).
For this, we employ a guided diffusion model, trained via a simple imitation objective on trajectory data to learn the underlying distribution of expert driving behavior from a large-scale dataset.
Following \texttt{GUMP}~\cite{hu2024solving}, \texttt{STR}~\cite{sun2025generalizing}, and \texttt{DiffusionPlanner}~\cite{zheng2025diffusion}, we choose a \texttt{DiT}-architecture~\cite{peebles2023scalable} as our backbone.
At each denoising step $t$, the trajectory score is steered by the learned data distribution and by differentiable driving objectives
\begin{equation}
\label{eq:objective_function}
    \mathcal{O}(\mathbf{x}_t) = \mathcal{O}_{\text{collision}}(\mathbf{x}_t) + \mathcal{O}_{\text{drivable}}(\mathbf{x}_t) + \mathcal{O}_{\text{progress}}(\mathbf{x}_t) + \mathcal{O}_{\text{comfort}}(\mathbf{x}_t).
\end{equation}
Here, $\mathbf{x}_t$ is the generated trajectory, and the objective terms penalize collision risk, deviation from the drivable area, lack of progress, and uncomfortable maneuvers.
The gradient $-\nabla \mathcal{O}(\mathbf{x}_t)$ implicitly guides denoising towards safe, efficient, and comfortable trajectories, yielding a single, scene-consistent joint trajectory proposal already biased towards satisfying our core driving principles.
Using one objective across modules is a deliberate trade-off: specialized modules can tune local behavior more strongly toward individual objectives, but may also create incompatible preferences between proposal generation and refinement.
We therefore use the shared objective as a conservative coordination mechanism, without claiming it is universally preferable to specialized modular designs.\\
\textbf{Collision.}
Following Zheng~\etal~\cite{zheng2025diffusion}, we penalize proximity to other agents using the signed distance $D$ between the ego bounding box and each agent $M$ at pose $\tau \in \mathbb{N}, \tau < \tau_\text{max}$ along trajectory $\mathbf{x}_t^{\tau}$.
Distances below sensitivity radius $r$ receive an exponential penalty shaped by $\Psi(D) := e^D - D$, with a stronger penalty for overlap $(D < 0)$.
$w_{collision}$ controls penalty weight and sensitivity, and $\epsilon$ avoids zero-division
\begin{multline}
\label{eq:guidance_collision}
    \mathcal{O}_{\text{collision}} = \frac{\sum_{M,\tau} \mathbb{1}_{D^{\tau}_M>0} \cdot \Psi(w_{\text{collision}} \cdot \max(1 - \frac{D^{\tau}_M}{r}, 0))}{\sum_{M,\tau} \mathbb{1}_{D^{\tau}_M>0} + \epsilon}\\
    + \frac{\sum_{M,\tau} \mathbb{1}_{D^{\tau}_M<0} \cdot \Psi(w_{\text{collision}} \cdot \max(1 - \frac{D^{\tau}_M}{r}, 0))}{\sum_{M,\tau} \mathbb{1}_{D^{\tau}_M<0} + \epsilon}.
\end{multline}
\\
\textbf{Drivable Area Compliance.}
We compute a differentiable cost map $C$ with a Euclidean Signed Distance Field~\cite{cheng2024pluto, zheng2025diffusion} and penalize any ego pose $\mathbf{x}_{t, \text{ego}}^\tau$ outside the drivable area
\begin{equation}
\label{eq:guidance_drivable}
    \mathcal{O}_{\text{drivable}} = \frac{\sum_{\tau} \Psi(w_{\text{drivable}} \cdot C(\mathbf{x}_{t, \text{ego}}^\tau))}{\sum_{\tau} \mathbb{1}_{C(\mathbf{x}_{t, \text{ego}}^\tau)>0} + \epsilon}.
\end{equation}
\\
\textbf{Progress.}
We encourage efficient progress by penalizing squared deviations from target speed $v_{\text{target}}$, but only outside tolerance $\delta$.
For desired speed range $[v_{\text{low}}, v_{\text{high}}]$ with $v_{\text{high}} \leq v_{\text{limit}}$, we set $v_{\text{target}}=\frac{v_{\text{low}}+v_{\text{high}}}{2}$ and $\delta=\frac{v_{\text{high}}-v_{\text{low}}}{2}$, yielding
\begin{equation}
\label{eq:guidance_progress}
\mathcal{O}_{\text{progress}} = w_{\text{progress}} \cdot \max\left( \left|\frac{\overline{d\mathbf{x}^{\tau}_{t, \text{ego}}}}{d\tau} - v_{\text{target}}\right| - \delta, 0 \right)^2.
\end{equation}
\\
\textbf{Comfort.}
We penalize excessive jerk, i.e., the change rate of acceleration, beyond longitudinal and lateral jerk thresholds $j_{\text{lon, max}}$ and $j_{\text{lat, max}}$
\begin{equation}
\label{eq:guidance_comfort}
\begin{split}
    \mathcal{O}_{\text{comfort}} = w_{\text{comfort}} \cdot \mathbb{E}_\tau \left[ \vphantom{\frac{1}{2} \max\left( \left( \left| j_{\text{lat,max}} - \frac{d^3y^{\tau}_{\text{ego}}}{d\tau^3} \right| \right) \Delta\tau^3, 0 \right)^2} \right. & \frac{1}{2} \max\left( \left( \left| j_{\text{lon,max}} - \frac{d^3\mathbf{x}^{\tau}_{t, \text{ego}}}{d\tau_\text{lon}^3} \right| \right) \Delta\tau_\text{lon}^3, 0 \right)^2 \\
    & \left. + \frac{1}{2} \max\left( \left( \left| j_{\text{lat,max}} - \frac{d^3\mathbf{x}^{\tau}_{t, \text{ego}}}{d_\text{lat}\tau^3} \right| \right) \Delta\tau_\text{lat}^3, 0 \right)^2 \right].
\end{split}
\end{equation}
Selecting $w_{\text{collision}} > w_{\text{drivable}} >> w_{\text{progress}} \approx w_{\text{comfort}}$ prioritizes safety over comfort for the generated initial ego trajectory.

\subsection{MCTS in Action Space}
While the diffusion model produces a globally plausible plan, it does not by itself provide explicit constraint checks required for safety-critical deployment.
The second stage of \texttt{DiffuSearch} is therefore an MCTS planner that performs an explicit, interpretable refinement of the initial proposal.
Following recent works in autonomous driving~\cite{chekroun2024mbappe, yu2025hype}, we define the search over a discretized action space composed of combined longitudinal (acceleration $\alpha$) and lateral (steering angle $\varphi$) command tuples $a\, \hat{=}\, (\alpha, \varphi)$.
This approach provides two key benefits: explainability and explicit constraint enforcement throughout the search steps (\cref{fig:model_overview}).\\
\textbf{Selection.}
Starting from the root node, which represents the current vehicle state, the algorithm recursively traverses the tree by selecting the child node that maximizes the PUCT (Polynomial Upper Confidence bound for Trees) criterion~\cite{rosin2011multi}.
This ensures a balance between exploiting high-value branches and exploring less-visited ones.
For a given state $s$, the next action $a$ is selected as
\begin{equation}
    a_t = \arg\max_{a} \left( Q(s, a) + c_{\text{puct}} \cdot P(s, a) \frac{\sqrt{\sum_b N(s, b)}}{1 + N(s, a)} \right).
\end{equation}
where $Q(s, a)$ is the mean action value (derived from reward~$\mathcal{R}$, cf{.} \cref{eq:mcts_backup}), $N(s, a)$ is the visit count, $c_{\text{puct}}$ is a hyperparameter balancing exploration, and $P(s, a)$ is the prior probability of selecting action $a$.
Unlike methods that use a broad prior~\cite{chekroun2024mbappe, yu2025hype}, we use the diffusion-generated ego trajectory to provide a strong, focused prior.
Specifically, $P(s, a)$ is modeled as a tight Gaussian distribution centered on the discrete action that is closest to the action prescribed by our diffusion proposal at that timestep.
This focuses the MCTS search effort on a local, scene-consistent region of the action space.\\
\textbf{Expansion.}
When selection reaches a leaf node, all feasible actions from the discretized set create child nodes.
Feasibility follows continuity constraints~\cite{chekroun2024mbappe} on acceleration and steering changes, promoting comfort, physical feasibility, and locality.\\
\textbf{Simulation.}
When expanding a leaf node, simulation updates the scene based on the action and evaluates the resulting state.
A kinematic bicycle model transfers the ego action $(\alpha, \varphi)$ into a new pose, promoting physically plausible ego motion under normal operating conditions~\cite{riekert1940fahrmechanik}.
As with any low-order vehicle model, this approximation can become inaccurate under extreme lateral acceleration or tire-slip regimes.
All other agents are updated using the joint predictions from our diffusion model at 10\,Hz.
This provides a realistic world model for the MCTS to plan against and is a reasonable assumption due to the locality of our refinement.
The value of a simulated trajectory is explicitly calculated as a reward $\mathcal{R}(\mathbf{x}_{\text{ego}})$ by summing the same four driving objectives used for the implicit diffusion guidance
\begin{equation}
    \mathcal{R}(\mathbf{x}_{\text{ego}}) = \mathcal{R}(\mathbf{x}_{\text{ego}})_{\text{collision}} + \mathcal{R}(\mathbf{x}_{\text{ego}})_{\text{drivable}} + \mathcal{R}(\mathbf{x}_{\text{ego}})_{\text{progress}} + \mathcal{R}(\mathbf{x}_{\text{ego}})_{\text{comfort}}.
\end{equation}
The guidance functions in \cref{eq:guidance_collision} to \cref{eq:guidance_comfort} are non-negative costs ${\mathcal{O} \in [0, +\infty)}$ with 0 as ideal outcome, so we use inverse rewards $\mathcal{R} \propto -\mathcal{O} \in (-\infty, 0]$.
We use the same weights $w_{\text{collision}}$, $w_{\text{drivable}}$, $w_{\text{progress}}$ and $w_{\text{comfort}}$ as for the guidance functions (\cref{sec:method_diffusion}), ensuring that MCTS optimizes the objectives that guided proposal generation.
\\
\textbf{Backpropagation.}
The simulation reward $\mathcal{R}$ is backpropagated to the root node, updating visit counts $N(s, a)$ and mean action values $Q(s, a)$ along the traversed search path
\begin{equation}
\label{eq:mcts_backup}
    N(s, a) \leftarrow N(s, a) + 1\,; \quad
    Q(s, a) \leftarrow \frac{N(s, a) \cdot Q(s, a) + \mathcal{R}(\mathbf{x}_{\text{ego}})}{N(s, a) + 1}.
\end{equation}
By repeatedly performing these steps, the MCTS builds a focused search tree in action space, using the diffusion proposal as its central prior.\\
\textbf{Final Trajectory Planning.}
After a fixed computational budget, e.g., 256 search steps, the optimal trajectory is retrieved from the tree.
Starting from root state $s_0$, we greedily follow the child with the highest visit count $N(s, a)$, yielding action sequence $A^\ast = (a_0^\ast, a_1^\ast, \dots, a_{\tau_\text{max}-1}^\ast)$, with $a^\ast_\tau \hat{=} (\alpha^\ast_\tau, \varphi^\ast_\tau)$, and state sequence $S^\ast = (s_0, s_1^\ast, \dots, s_{\tau_\text{max}}^\ast)$
\begin{equation}
    a_\tau^* = \underset{a \in \mathcal{A}(s_\tau^*)}{\arg\max} \, N(s_\tau^*, a)
\end{equation}
where $\mathcal{A}(s_\tau^\ast)$ is the set of all possible actions from state $s_\tau^\ast$.
A kinematic bicycle model transforms this action sequence into the final ego trajectory.
Since the plan remains contingent on the predicted scene evolution, we include the emergency braking mechanism used by most hybrid planners~\cite{vitelli2022safetynet, phan2023driveirl, dauner2023parting, sun2025generalizing, zheng2025diffusion}.
However, this is scarcely needed since at 10\,Hz replanning, \texttt{DiffuSearch} quickly adapts to atypical scene development in the next cycle.

\subsection{Experimental Setup}
\textbf{Datasets, Benchmarks \& Metrics.}
We develop our \texttt{DiffuSearch} model in the nuPlan framework~\cite{karnchanachari2024towards}, which contains $\sim$1{,}300 hours real world driving data. 
Since many works have shown that open-loop evaluation weakly correlates with real world driving behavior~\cite{codevilla2018offline, dauner2023parting, hagedorn2025planners}, we focus on reactive closed-loop simulation.
nuPlan’s closed-loop simulator runs each scenario for 15\,s at 10\,Hz, starting from a real traffic situation.
While the ego vehicle is controlled by the planner and a low-level controller, other traffic agents are updated with a rule-based Intelligent Driver Model (\texttt{IDM})~\cite{treiber2000congested}.
Alternatively, learned, reactive \texttt{SMART} agents provide versatile behavior to test more complex interactions in closed-loop simulation~\cite{wu2024smart, hagedorn2025planners}.
On nuPlan, we use these two closed-loop benchmarks to measure the reactive R-score and the \texttt{SMART}-reactive SR-score.
Both scores are normalized to $[0, 100]$ and are derived from the number of collisions (NC), time-to-collision (TTC), drivable area compliance (DAC), progress, and comfort~\cite{karnchanachari2024towards, dauner2023parting}.
Results are reported on three data splits: Val14, Test14, and Test14-hard, which contain 1118, 280, and 272 traffic scenarios, respectively.
Test14-hard is a curated data split of difficult scenarios with a high degree of interaction between traffic participants like unprotected turns or tight merges~\cite{cheng2024rethinking}, and hence suitable to differentiate the capability of planners with similar performance on the standard Val14 split.
To test cross-benchmark transfer beyond nuPlan, we additionally use the interPlan lane-change benchmark with low-, medium-, and high-density traffic settings~\cite{hallgarten2024can}.\\
\textbf{Baseline Models.}
Our model is compared against the best other hybrid planners in nuPlan.
\texttt{DTPP} and \texttt{DiffusionPlanner} are the most similar to our work, as they apply a learned tree search, and a diffusion backbone, respectively (see \cref{sec:methodology}).
To facilitate a clear investigation on the effect of MCTS-based trajectory refinement and aligned objectives in diffusion guidance and MCTS reward, we use the same pretrained diffusion backbone as \texttt{DiffusionPlanner} for our experiments with \texttt{DiffuSearch}.

\section{Experiments \& Results}
Comprehensive quantitative and qualitative results of our \texttt{DiffuSearch} planner are presented in this section.
We structure the analysis around the central insights gained on the nuPlan splits and the interPlan lane-change benchmark.

\begin{figure}[!tb]
    \centering
    \includegraphics[width=\columnwidth]{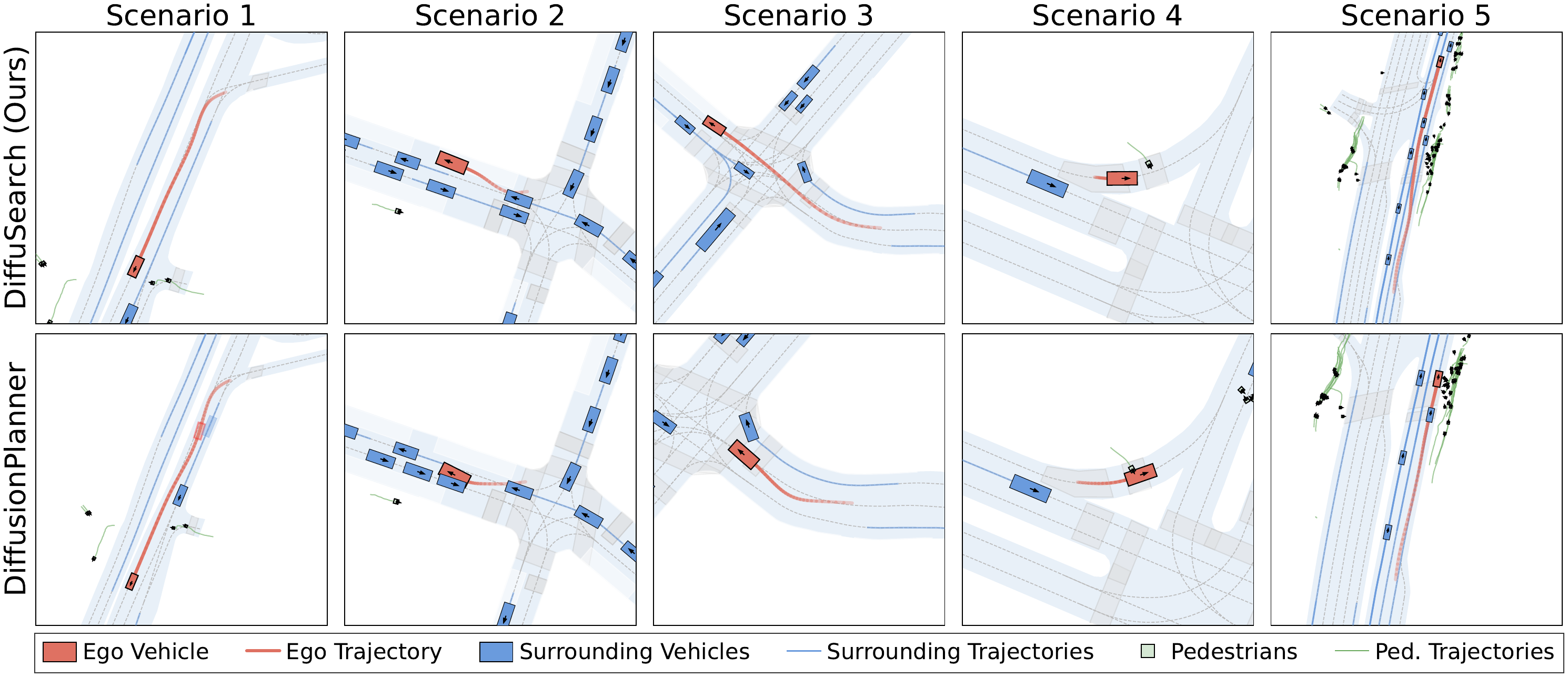}
    \caption{\textbf{Closed-loop behavior} of \texttt{DiffuSearch} (top row) compared to the strongest baseline \texttt{DiffusionPlanner} (bottom row) on five nuPlan Test14-hard-R scenarios. \texttt{DiffuSearch} corrects collision-prone diffusion proposals, keeps larger safety margins, waits for pedestrians, and proactively changes lanes in critical situations.}
    \label{fig:qualitative_results}
\end{figure}

\subsection{DiffuSearch excels on hard, interactive scenarios.}
\begin{table*}[!tb]
    \centering
    \caption{\textbf{Planner evaluation on nuPlan reactive closed-loop benchmarks.} R means closed-loop simulation with nuPlan's reactive \texttt{IDM} agents~\cite{treiber2000congested}. SR is measured using \texttt{SMART}-reactive agents~\cite{wu2024smart, hagedorn2025planners}. Bold indicates the best-performing method, and underline indicates the second-best.}
    \small
    \begin{tabular*}{1.0\textwidth}{@{\extracolsep{\fill}} l r @{\quad} r @{\qquad} r @{\quad} r @{\qquad} r @{\quad} r}
        \toprule
        \multicolumn{1}{c}{\textbf{Planner}} & \multicolumn{2}{c}{\textbf{Val14}} & \multicolumn{2}{c}{\textbf{Test14}} & \multicolumn{2}{c}{\textbf{Test14-hard}} \\
        \cmidrule(lr){1-1}\cmidrule(lr){2-3}\cmidrule(lr){4-5}\cmidrule(lr){6-7}
        \textbf{} & \textbf{R} $\uparrow$ & \textbf{SR} $\uparrow$ & \textbf{R} $\uparrow$ & \textbf{SR} $\uparrow$ & \textbf{R} $\uparrow$ & \textbf{SR} $\uparrow$ \\
        \midrule
        \texttt{PDM-Hybrid}~\cite{dauner2023parting}  & \textbf{92.11} & \underline{89.44} & 91.28 & \underline{90.24} & 76.07 & 72.15 \\
        \texttt{GameFormer}~\cite{huang2023gameformer}  & 79.78 & 78.05 & 82.05 & 79.33 & 67.05 & 62.48 \\
        \texttt{DTPP}~\cite{huang2024dtpp}  & 63.95 & 62.41 & 73.88 & 64.82 & 47.47 & 54.65 \\
        \texttt{PLUTO}~\cite{cheng2024pluto}  & 90.48 & 83.88 & 90.29 & 85.27 & 76.88 & 69.36 \\
        \texttt{DiffusionPlanner}~\cite{zheng2025diffusion}  & 91.33 & 88.84 & \underline{92.28} & 88.97 & \underline{78.79} & \underline{75.25} \\
        \texttt{DiffuSearch} (ours)  & \underline{91.39} & \textbf{89.55} & \textbf{93.03} & \textbf{91.12} & \textbf{79.98} & \textbf{77.07} \\
        \bottomrule
    \end{tabular*}
    \label{tab:main_results}
\end{table*}
A comparison of \texttt{DiffuSearch} with the best hybrid planners in reactive (R) and \texttt{SMART}-reactive (SR) closed-loop simulation on the three nuPlan data splits is shown in \cref{tab:main_results}.
\texttt{DiffuSearch} achieves state-of-the-art results on all SR benchmarks and two out of three R benchmarks.
Compared to other models, our method performs especially well in hard, interactive traffic scenarios.
Specifically, our method generates an improvement of up to $1.82\%$ compared to the strong \texttt{DiffusionPlanner} baseline and up to $4.92\%$ over the nuPlan challenge winner \texttt{PDM-Hybrid} on Test14-hard-SR.
On Val14-R, \texttt{DiffuSearch} lies slightly behind the \texttt{PDM-Hybrid} baseline.
This could be explained by the strong centerline-following prior of this baseline.
The standard driving scenarios of Val14 together with the cautious behavior of uniformly lane-following \texttt{IDM} agents in reactive closed-loop simulation lets \texttt{PDM-Hybrid} naturally blend in.
However, as soon as interactive \texttt{SMART} agents challenge the planner with more realistic and diverse behaviors, our method outperforms the baseline.
When increasing the difficulty further and switching to the harder scenarios in Test14-hard, the generative backbone and the MCTS refinement let \texttt{DiffuSearch} find better solutions for complex situations.
\cref{fig:qualitative_results} shows a qualitative comparison of our model with the strongest baseline on this data split.
Since both models use the same, pretrained diffusion backbone, the difference is consistent with the effect of applying guidance and MCTS refinement.
\cref{fig:qualitative_results} illustrates the type of local repair targeted by our refiner: \texttt{DiffuSearch} steers diffusion-based trajectories away from collision-prone interactions with other vehicles and pedestrians, smooths abrupt maneuvers, and even changes the lane after anticipating a critical situation with pedestrians walking at the edge of the road.
To test whether this advantage transfers beyond the nuPlan splits, we additionally evaluate both planners on the interPlan lane-change benchmark, which stresses repeated lane changes across dense reactive traffic~\cite{hallgarten2024can}.
The results in \cref{tab:interplan_results} show that \texttt{DiffuSearch} consistently outperforms \texttt{DiffusionPlanner} across all traffic densities and under both \texttt{IDM} and \texttt{SMART} background traffic.
The gains are particularly pronounced in medium- and high-density settings, where safe progress depends on repeated local corrections and robust interaction handling.
This supports our central claim that aligned guidance and explicit local search are most beneficial when the planner must negotiate tightly coupled maneuvers rather than merely follow the lane.

\begin{table}[!tb]
    \centering
    \caption{\textbf{Cross-benchmark evaluation on the interPlan lane-change benchmark}~\cite{hallgarten2024can}. Reactive closed-loop scores are reported on the benchmark's native $[0,1]$ scale across three traffic densities.}
    \small
    \setlength{\tabcolsep}{3.2pt}
    \begin{tabular*}{0.8\linewidth}{@{\extracolsep{\fill}}lcccccc@{}}
        \toprule
        \multirow{2}{*}{\textbf{Planner}} & \multicolumn{3}{c}{\textbf{\texttt{IDM}}} & \multicolumn{3}{c}{\textbf{\texttt{SMART}}} \\
        \cmidrule(lr){2-4}\cmidrule(lr){5-7}
        & \textbf{Low} & \textbf{Med.} & \textbf{High} & \textbf{Low} & \textbf{Med.} & \textbf{High} \\
        \midrule
        \texttt{DiffusionPlanner}~\cite{zheng2025diffusion} & 0.642 & 0.583 & 0.598 & 0.583 & 0.532 & 0.350 \\
        \texttt{DiffuSearch} (ours)      & \textbf{0.688} & \textbf{0.679} & \textbf{0.629} & \textbf{0.630} & \textbf{0.578} & \textbf{0.419} \\
        \bottomrule
    \end{tabular*}
    \label{tab:interplan_results}
\end{table}

\subsection{MCTS refinement consistently improves results.}
\begin{table*}[!tb]
    \centering
    \caption{\textbf{Ablation of guidance and MCTS refinement.} Guidance alone degrades scores, MCTS alone improves them, and their aligned combination performs best.}
    \resizebox{1.0\textwidth}{!}{
    \begin{tabular}{l ccc r @{\qquad} r @{\quad} r @{\qquad} r @{\quad} r @{\qquad} r}
        \toprule
        \multicolumn{1}{c}{\textbf{Planner}} & \multicolumn{3}{c}{\textbf{Module}} & \multicolumn{2}{c}{\textbf{Val14}} & \multicolumn{2}{c}{\textbf{Test14}} & \multicolumn{2}{c}{\textbf{Test14-hard}} \\
        \cmidrule(lr){1-1}\cmidrule(lr){2-4}\cmidrule(lr){5-6}\cmidrule(lr){7-8}\cmidrule(lr){9-10}
        \textbf{} & \textbf{Diffusion} & \textbf{Guidance} & \textbf{MCTS} & \textbf{R} $\uparrow$ & \textbf{SR} $\uparrow$ & \textbf{R} $\uparrow$ & \textbf{SR} $\uparrow$ & \textbf{R} $\uparrow$ & \textbf{SR} $\uparrow$ \\
        \midrule
        \texttt{DiffusionPlanner~\cite{zheng2025diffusion}} & \checkmark & & & 91.33 & 88.84 & 92.28 & 88.97 & 78.79 & 75.25 \\
        \texttt{DiffuSearch} (ours) & \checkmark & \checkmark & & 89.88 & 86.77 & 91.31 & 86.87 & 77.88 & 74.41 \\
        \texttt{DiffuSearch} (ours) & \checkmark & & \checkmark & \underline{91.36} & \underline{89.32} & \underline{92.59} & \underline{89.84} & \underline{79.52} & \underline{76.36} \\
        \texttt{DiffuSearch} (ours) & \checkmark & \checkmark & \checkmark & \textbf{91.39} & \textbf{89.55} & \textbf{93.03} & \textbf{91.12} & \textbf{79.98} & \textbf{77.07} \\
        \bottomrule
    \end{tabular}
    }
    \label{tab:ablation_guidance_mcts}
\end{table*}
In the ablation experiments shown in \cref{tab:ablation_guidance_mcts}, we investigate the influence of classifier-free diffusion guidance when generating the initial joint prediction and the subsequent MCTS refinement of the ego trajectory in action space.
The results are consistent across all data splits and benchmarks: adding guidance alone does not improve planning performance.
More severely, it even decreases the score by up to $-1.45\%$ on Val14-R and $-2.1\%$ on Test14-SR, which might explain why no quantitative results have been published so far.
In contrast, adding MCTS refinement alone consistently leads to improvements of up to $+1.11\%$ on Test14-hard-SR and never decreases planning performance.
Finally, when combining diffusion guidance and MCTS refinement in action space, the best performance is achieved on all data splits and benchmarks.
This is an unexpected result since diffusion guidance alone decreases performance.
However, when combined with MCTS refinement, performance increases beyond simply adding up the impact of the individual design choices of diffusion and MCTS refinement.
We hypothesize that this could be related to guidance function design.
Zheng~\etal~\cite{zheng2025diffusion} report that classifier-free guidance delicately depends on the guidance functions' properties.
They state that guidance functions should ideally be sparse, i.e., only influence denoising in critical situations, address higher-order state derivatives indirectly, and create smooth and continuous gradients of consistent magnitude.
Combining the learned denoising gradient $s_\theta(x_t, t)$ with a hand-crafted objectives gradient $\nabla_{x_t} \mathcal{O}(x_t, t)$ is susceptible to emergent effects, where the diffusion model is pushed into regions outside of the distribution encountered during training without guidance.
This could locally optimize the generated trajectory w{.}r{.}t{.} the guidance objectives but result in globally inconsistent solutions.
Adding MCTS refinement smooths the inconsistencies caused by the described emerging effects and even seems to benefit from the local ``hints'' provided through guidance.
This interpretation is corroborated by the trajectory-consistency analysis in \cref{tab:trajectory_consistency}.
Compared to \texttt{PLUTO}, which also applies rule-based post-processing to a learned proposal~\cite{cheng2024pluto}, \texttt{DiffuSearch} yields substantially smaller deviations between the initial and refined trajectories in terms of position and speed while maintaining similarly small heading corrections.
The refiner therefore acts as a local repair mechanism rather than replacing the diffusion proposal with a qualitatively different plan.
This is the behavior intended with the unified objectives: the generator already places the plan in a promising region of the solution space, and MCTS refines the same objectives locally.
In summary, the ablation results suggest that aligning the objectives across the model synchronizes trajectory generation and refinement, leading to a more consistent planning process.

\begin{table*}[!tb]
    \centering
    \caption[Trajectory consistency between the initial and refined trajectory]{\textbf{Trajectory consistency} between the initial and refined trajectory on nuPlan Test14. Lower values indicate that the refinement stays closer to the original proposal.}
    \resizebox{0.75\textwidth}{!}{
    \begin{tabular}{llccc}
        \toprule
        \textbf{Agents} & \textbf{Planner} & \textbf{$\Delta$ L2 [m]} & \textbf{$\Delta$ Heading [rad]} & \textbf{$\Delta$ Speed [m/s]} \\
        \midrule
        \multirow{2}{*}{\texttt{IDM}}   & \texttt{PLUTO}~\cite{cheng2024pluto}       & 1.469 $\pm$ 1.410          & \textbf{0.015 $\pm$ 0.012} & 0.474 $\pm$ 0.432 \\
                                         & \texttt{DiffuSearch} (ours) & \textbf{0.754 $\pm$ 0.484} & 0.016 $\pm$ 0.020          & \textbf{0.382 $\pm$ 0.381} \\
        \midrule
        \multirow{2}{*}{\texttt{SMART}} & \texttt{PLUTO}~\cite{cheng2024pluto}       & 3.373 $\pm$ 2.713          & 0.030 $\pm$ 0.031          & 1.061 $\pm$ 0.804 \\
                                         & \texttt{DiffuSearch} (ours) & \textbf{1.859 $\pm$ 1.810} & \textbf{0.022 $\pm$ 0.028} & \textbf{0.919 $\pm$ 1.241} \\
        \bottomrule
    \end{tabular}}
    \label{tab:trajectory_consistency}
\end{table*}

\begin{figure}[!tb]
    \centering
    \includegraphics[width=0.9\linewidth]{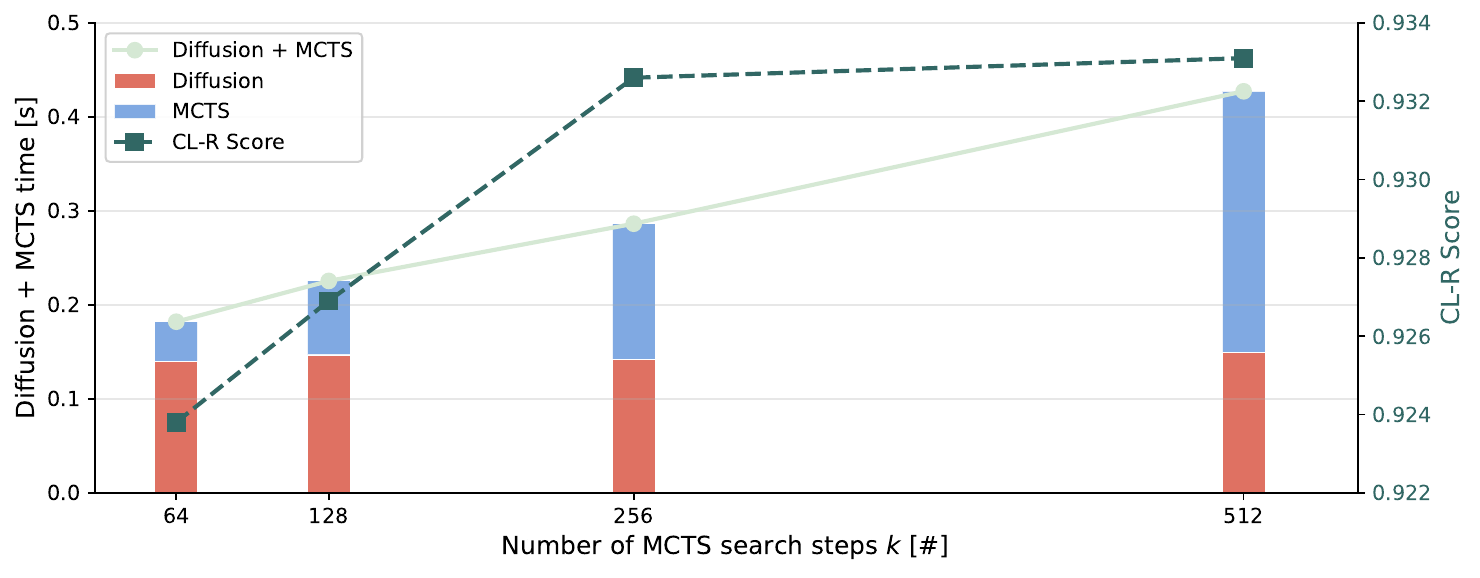}
    \caption{\textbf{Runtime and performance vs. MCTS search steps.} Diffusion dominates runtime for small budgets, while MCTS adds linearly with the number of search steps and improves performance until local refinement saturates around 256 steps. Runtime is measured on a Python implementation without runtime optimizations. The CL-R score axis is zoomed to make the differences better visible.}
    \label{fig:runtime_vs_k}
\end{figure}

The additional search cost of this local refinement is analyzed in \cref{fig:runtime_vs_k}.
For small search budgets, the majority of the runtime is spent in the diffusion backbone, while the MCTS contribution grows approximately linearly with the number of search steps.
At the same time, planning performance improves steadily until the local refinement converges around 256 search steps, which is therefore the operating point used throughout our experiments.
This trade-off highlights that the explicit search is not used for broad global exploration, but for focused local improvement around a strong learned prior.

\subsection{Unified objectives provide additional gains.}
\begin{table*}[!tb]
    \centering
    \caption{\textbf{Ablation study of guidance and reward components.} The impact of individual objective terms for both guidance and MCTS reward is evaluated on the Test14-hard benchmark. The final row represents the full model configuration.}
    \resizebox{1.0\textwidth}{!}{
    \begin{tabular}{cccc @{\qquad} cccc @{\qquad} r}
        \toprule
        \multicolumn{4}{c}{\textbf{Guidance Objective}} & \multicolumn{4}{c}{\textbf{MCTS Reward}} & \multicolumn{1}{c}{\textbf{Test14-hard}} \\
        \cmidrule(lr){1-4}\cmidrule(lr){5-8}\cmidrule(lr){9-9}
        \textbf{Collision} & \textbf{Drivable} & \textbf{Progress} & \textbf{Comfort} & \textbf{Collision} & \textbf{Drivable} & \textbf{Progress} & \textbf{Comfort} & \textbf{R} $\uparrow$ \\
        \midrule
         & \checkmark & \checkmark & \checkmark & \checkmark & \checkmark & \checkmark & \checkmark & 79.44 \\ 
        \checkmark &  & \checkmark & \checkmark & \checkmark & \checkmark & \checkmark & \checkmark & 78.77 \\ 
        \checkmark & \checkmark &  & \checkmark & \checkmark & \checkmark & \checkmark & \checkmark & 79.18 \\ 
        \checkmark & \checkmark & \checkmark &  & \checkmark & \checkmark & \checkmark & \checkmark & 79.01 \\ 
        \checkmark & \checkmark & \checkmark & \checkmark &  & \checkmark & \checkmark & \checkmark & 42.53 \\ 
        \checkmark & \checkmark & \checkmark & \checkmark & \checkmark &  & \checkmark & \checkmark & 72.58 \\ 
        \checkmark & \checkmark & \checkmark & \checkmark & \checkmark & \checkmark &  & \checkmark & 66.24 \\ 
        \checkmark & \checkmark & \checkmark & \checkmark & \checkmark & \checkmark & \checkmark &  & 79.34 \\
        \hline
        \checkmark & \checkmark & \checkmark & \checkmark & \checkmark & \checkmark & \checkmark & \checkmark & \textbf{79.98} \\
        \bottomrule
    \end{tabular}
    }
    \label{tab:ablation_individual_objectives}
\end{table*}
The previous experiments show that guidance and MCTS refinement together consistently yield the best planning performance.
In a second ablation study we switch off individual components of the guidance objective~$\mathcal{O}$ and the reward~$R$ (cf. \cref{sec:methodology}).
The results are shown in \cref{tab:ablation_individual_objectives}.
In summary, each component contributes to the overall performance and using the same objective set in diffusion guidance and MCTS reward yields the best observed planned trajectory.
Nevertheless, there are large differences in the importance of the individual components.
Generally, switching off guidance objectives leads to much smaller performance decreases than switching off MCTS rewards.
This is expected, when considering the role these objectives play in the two modules.
Diffusion guidance objectives are designed to only affect the generated trajectory when constraints are violated.
They selectively adapt the imitation-based joint prediction that serves as the initial ego plan.
Conversely, the MCTS shapes the final plan based on the defined reward, which directly penalizes violations during refinement.
Ablating a component of the reward means removing this term from the optimization problem, so that the final trajectory is no longer directly penalized for violating this objective.
An apparent outlier is the comfort reward, which hardly changes the planner's behavior.
This result is grounded in the kinematic bicycle model used in our MCTS refinement.
It promotes that the planned trajectory already is smooth and feasible for the vehicle without the comfort reward.
Altogether, our result suggests that unified objectives in diffusion guidance and MCTS reward provide additional gains for trajectory planning.

\subsection{DiffuSearch reduces collisions and boosts comfort.}
\begin{table*}[!tb]
    \centering
    \caption{\textbf{Metric comparison} against \texttt{DiffusionPlanner} with \texttt{SMART} agents. \texttt{DiffuSearch} improves NC, TTC, and comfort while maintaining competitive DAC and progress.}
    \resizebox{1.0\textwidth}{!}{
    \begin{tabular}{l @{\quad} l @{\quad} r @{\quad} r @{\quad} r @{\quad} r @{\quad} r @{\quad} r}
        \toprule
        \textbf{Split} & \textbf{Planner} & \textbf{Score} $\uparrow$ & \textbf{NC} $\uparrow$ & \textbf{TTC} $\uparrow$ & \textbf{DAC} $\uparrow$ & \textbf{Comfort} $\uparrow$ & \textbf{Progress} $\uparrow$ \\
        \midrule
        \multirow{2}{*}{Val14}
        & \texttt{DiffusionPlanner~\cite{zheng2025diffusion}} & 88.84 & 97.83 & 92.41 & \textbf{99.82} & 70.34 & \textbf{96.93} \\
        & \texttt{DiffuSearch (ours)}      & \textbf{89.55} & \textbf{98.92} & \textbf{94.03} & 99.37 & \textbf{86.53} & 96.38 \\
        \midrule
        \multirow{2}{*}{Test14}
        & \texttt{DiffusionPlanner~\cite{zheng2025diffusion}} & 88.97 & 97.83 & 94.22 & 99.28 & 72.20 & 96.03 \\
        & \texttt{DiffuSearch (ours)}      & \textbf{91.12} & \textbf{98.19} & \textbf{95.67} & 99.28 & \textbf{87.73} & 96.03 \\
        \midrule
        \multirow{2}{*}{Test14-hard}
        & \texttt{DiffusionPlanner~\cite{zheng2025diffusion}} & 75.25 & 96.63 & 90.26 & \textbf{98.50} & 65.92 & \textbf{90.26} \\
        & \texttt{DiffuSearch (ours)}      & \textbf{79.98} & \textbf{98.50} & \textbf{94.01} & 97.00 & \textbf{83.90} & 88.76 \\
        \bottomrule
    \end{tabular}
    }
    \label{tab:metric_comparison}
\end{table*}

Finally, we conduct a detailed study of individual driving metrics to understand how our model behaves differently compared to the strongest baseline to achieve state-of-the-art scores on various nuPlan benchmarks.
Since \texttt{DiffuSearch} excels in realistic, interactive traffic scenarios, we perform this study on SR benchmarks.
We provide a breakdown of the individual driving metrics in \cref{tab:metric_comparison}.
Overall, the results are consistent with the priority of driving objectives we formulated in \cref{sec:methodology}, namely $w_{\text{collision}} > w_{\text{drivable}} >> w_{\text{progress}} \approx w_{\text{comfort}}$.
The safety-critical and most relevant objective of collision avoidance (NC) improves as the number of collisions reduces.
Moreover, \texttt{DiffuSearch} increases the time-to-collision (TTC), resulting in larger safety margins to adjacent vehicles, pedestrians, and bicycles, as exemplarily shown in \cref{fig:qualitative_results}.
This improvement in collision avoidance results in unchanged or slightly decreased drivable area compliance (DAC) and progress.
The planner prioritizes collision avoidance over drivable area compliance and, in critical scenarios, prevents collisions by swerving into parking bays and onto the shoulder (\cref{fig:qualitative_results}, scenario 2).
Due to the less offensive driving behavior showcased in scenario 3 of \cref{fig:qualitative_results}, our model progresses slightly slower compared to the strongest baseline.
This behavior is particularly acceptable as it serves the primary objective of collision avoidance.
Finally, though least important, \texttt{DiffuSearch} considerably improves ride comfort due to the kinematic model used in MCTS refinement by up to $+18\%$.

\section{Conclusion \& Future Work}
In this paper, we introduced \texttt{DiffuSearch}, a novel hybrid planner that synergistically combines a guided diffusion model with an MCTS refiner.
Our core contribution is the use of a unified set of driving objectives, which are used as implicit guidance for diffusion and as explicit rewards for MCTS.
Experiments on nuPlan and interPlan closed-loop benchmarks show that the combination of diffusion and an MCTS tightly focused around a learned prior already achieves strong performance, while sharing objectives across both stages provides additional improvements, consistently reducing collisions and improving ride comfort, particularly in challenging, interactive scenarios.
Future work will focus on two main extensions.
First, we will incorporate an error feedback mechanism, allowing the prediction model to learn from past errors and improve its temporal consistency during closed-loop inference.
Second, we aim to extend \texttt{DiffuSearch} to handle multi-modal proposals, enabling the MCTS to explicitly reason over distinct strategic maneuvers in ambiguous scenarios.
We believe these extensions, built upon the strong performance of our hybrid two-stage planner, are key steps toward creating even more robust and intelligent autonomous driving systems.

\newpage
%
%
\bibliographystyle{splncs04}
\bibliography{main}
\end{document}